\documentclass[twoside,leqno,twocolumn]{article}

\usepackage[letterpaper]{geometry}
\usepackage[inline]{enumitem}
\usepackage{ltexpprt}
\usepackage{hyperref}
\usepackage{tabu}
\usepackage{bm}
\usepackage{graphicx}
\usepackage{amsmath}
\usepackage{amssymb}
\usepackage{amsfonts}
\usepackage{latexsym}
\usepackage{verbatim}
\usepackage{hyperref}
\hypersetup{
	colorlinks   = true,
	citecolor    = blue,
	linkcolor    = blue,
	urlcolor     = blue,
}
\usepackage{cleveref}
\Crefname{equation}{Eq.}{Eqs.}
\usepackage{multirow}
\usepackage{geometry}
\usepackage[font=normal]{subcaption}
\usepackage[font=normal]{caption}
\usepackage[justification=centering]{caption}
\usepackage{indentfirst} 
\usepackage{filecontents}
\usepackage{tabu}
\usepackage{tabularx}
\newcommand{\tabincell}[2]{\begin{tabular}{@{}#1@{}}#2\end{tabular}}
\usepackage[dvipsnames]{xcolor}
\usepackage{ulem}
\usepackage{colortbl}
\usepackage{amsmath}
\usepackage{algorithm}
\usepackage{algpseudocode}
\usepackage[T1]{fontenc}
\begin{document}

\newcommand\relatedversion{}
\renewcommand\relatedversion{\thanks{The full version of the paper can be accessed at \protect\url{https://arxiv.org/abs/1902.09310}}} % Replace URL with link to full paper or comment out this line

%\setcounter{chapter}{2} % If you are doing your chapter as chapter one,
%\setcounter{section}{3} % comment these two lines out.

% \title{\Large SIAM/ACM Preprint Series Macros for Use With LaTeX\relatedversion}
% \author{Corey Gray\thanks{Society for Industrial and Applied Mathematics.}
% \and Tricia Manning\thanks{Society for Industrial and Applied Mathematics.}}

\title{\Large Deep Representation Learning for Multi-functional Degradation Modeling of Community-dwelling Aging Population}
\author{Suiyao Chen \thanks{\footnotesize Corresponding Author, sc3740@columbia.edu} \and
        Xinyi Liu  \and
        Yulei Li  \and
        Jing Wu \and 
        % Anna Hovakimyan \and 
        Handong Yao}

\date{}

\maketitle
\graphicspath{{./figures/figures/}}

% Copyright Statement
% When submitting your final paper to a SIAM proceedings, it is requested that you include
% the appropriate copyright in the footer of the paper.  The copyright added should be
% consistent with the copyright selected on the copyright form submitted with the paper.
% Please note that "20XX" should be changed to the year of the meeting.

% Default Copyright Statement
% \fancyfoot[R]{\scriptsize{Copyright \textcopyright\ 20XX by SIAM\\
% Unauthorized reproduction of this article is prohibited}}

% Depending on which copyright you agree to when you sign the copyright form, the copyright
% can be changed to one of the following after commenting out the default copyright statement
% above.

\fancyfoot[R]{\scriptsize{Copyright \textcopyright\ 2024\\
Copyright for this paper is retained by authors}}

%\fancyfoot[R]{\scriptsize{Copyright \textcopyright\ 20XX\\
%Copyright retained by principal author's organization}}

%\pagenumbering{arabic}
%\setcounter{page}{1}%Leave this line commented out.

\begin{abstract} \small\baselineskip=9pt 
As the aging population grows, particularly for the baby boomer generation, the United States is witnessing a significant increase in the elderly population experiencing multifunctional disabilities. These disabilities, stemming from a variety of chronic diseases, injuries, and impairments, present a complex challenge due to their multidimensional nature, encompassing both physical and cognitive aspects. Traditional methods often use univariate regression-based methods to model and predict single degradation conditions and assume population homogeneity, which is inadequate to address the complexity and diversity of aging-related degradation. This study introduces a novel framework for multi-functional degradation modeling that captures the multidimensional (e.g., physical and cognitive) and heterogeneous nature of elderly disabilities. Utilizing deep learning, our approach predicts health degradation scores and uncovers latent heterogeneity from elderly health histories, offering both efficient estimation and explainable insights into the diverse effects and causes of aging-related degradation. A real-case study demonstrates the effectiveness and marks a pivotal contribution to accurately modeling the intricate dynamics of elderly degradation, and addresses the healthcare challenges in the aging population.   
\end{abstract}

\section{Introduction}
The demographic evolution marked by the aging of the baby boomer generation heralds a forthcoming surge in the elderly population with disabilities. In 2008–2012, there were about 15.6 million disabled elderly people, which accounted for 38.7\% of the total number of U.S. older adults \cite{he2014older}. This underscores a pressing need in the healthcare industry to adequately prepare for an upswing in demand of health resources utilization and care services customization \cite{dall2013aging,chen2017personalized, lai2024language,yan2024self, read2023prediction}. Such disabilities manifest in multifaceted forms, arising from a combination of chronic diseases, injuries, and impairments that result from both physical and cognitive degradation \cite{demanze2017fall,chen2019claims}. 

Traditional approaches to examining age-related degradation have largely depended on univariate analyses, which simplistically view the elderly population as a homogeneous group \cite{kim2018unmet}. This type of simplification could work for various industries and applications such as manufacturing \cite{wu2023genco,chen2017multi,chen2020optimal,chen2020some,bingjie2023optimal,gao2023autonomous}, agriculture \cite{wu2022optimizing,tao2022optimizing,wu2023extended,wu2024new} and various engineering fields \cite{shi2017combining, chen2020some, wang2023inverse,liu2024particle,zhang2020manipulator}. However, this oversimplification may overlook the complex, multifaceted nature of aging, failing to account for its multidimensional and heterogeneous characteristics \cite{chen2018data, wang2017sensitivity, wang2019inverse, wang2023embracing, zhu2021taming}. For example, in the long-term care (LTC) system, the diversified needs of the elderly in nursing homes and assisted living facilities can be largely determined by their functional performance or disability and mental capability, both of which result from aging degradation and contribute to the overall health condition. 

To fill this gap and systematically investigate the heterogeneous performance degradation of the aging population, our research introduces an innovative multifunctional degradation modeling framework, uniquely designed to encapsulate the complex, multidimensional degradations—encompassing both physical and cognitive dimensions—that characterize the aging process. By leveraging a deep learning framework, our study transcends the traditional confines associated with predetermined homogeneous population, thus offering a more nuanced portrayal of aging's heterogeneous nature. This methodology not only augments our comprehension of aging dynamics but also refines the efficiency of healthcare utilization management.

This work is further distinguished by its application of cutting-edge deep Learning techniques \cite{wu2024switchtab,chen2023recontab,wang2023balanced,wu2023hallucination,lai2024adaptive, su2024large, tang2024fedlion,sun2018relation,li2024enhancing} to analyze longitudinal data derived from the Health and Retirement Study (HRS). This approach empowers us to unveil sophisticated representations of the survey data, illuminating the intricate patterns of aging-related degradation with unprecedented depth and accuracy. Through the application of these advanced analytical tools, we aim to dissect the multifarious impacts of aging on different healthcare settings, including nursing homes and hospitals, thereby furnishing critical insights to optimize resource allocation and care program designs. Moreover, our exploration extends to the etiological underpinnings of degradation heterogeneity, endeavoring to unravel the factors that precipitate variations in aging trajectories. Through the case study grounded in real-world HRS survey data, our research elucidates the practicality and robustness of our proposed modeling approach. This empirical investigation not only validates our methodology but also showcases its capability to inform policy decisions and healthcare practices tailored to the increasingly complex needs of the aging society.

The contributions are summarized as follows: 
\begin{itemize}[noitemsep,topsep=0pt]
    \item [$\bullet$] We proposed a deep learning framework for multi-functional degradation modeling and apply for aging population health prediction.

    \item [$\bullet$] We explicitly quantified the heterogeneity in the aging population and visualized the distinctive degradation and healthcare utilization patterns for each subpopulation. 

    \item [$\bullet$] We conducted a comprehensive empirical study on HRS survey data and demonstrated the superiority of the proposed work.

\end{itemize}

\section{Related Work}
\subsection{Data Science in Healthcare Data Analytics}
The integration of data science and healthcare analytics has been a pivotal advancement in enhancing healthcare delivery, patient care, and operational efficiency. Simpao et al.\cite{simpao2014review} highlight the significant role of analytics in leveraging the vast amount of patient data collected through electronic health record systems for predictive risk assessment and clinical decision support. Miotto et al. \cite{miotto2018deep} emphasize the potential of deep learning technologies in transforming healthcare by enabling end-to-end learning models from complex biomedical data. The categorization of big data uses in healthcare into administration, clinical decision support, and consumer behavior \cite{hermon2014big} underscores the versatile applications of analytics in improving patient outcomes and controlling healthcare costs. Galetsi and Katsaliaki \cite{galetsi2020review} provide an overview of big data analytics (BDA) publication dynamics, focusing on the application of modeling and machine learning techniques for health monitoring and prediction. Collectively, these studies illustrate the transformative potential and challenges of applying data science in healthcare, emphasizing the necessity for interdisciplinary expertise and privacy-conscious approaches in exploiting the full potential of big data analytics.
\subsection{Aging and Health Research}
In the intersection of aging, health, and data science, recent studies have begun to harness big data and machine learning to explore and address the complexities of aging. Zhang et al.\cite{zhang2021role} propose an ecological framework for utilizing big data in aging research, emphasizing the need for interdisciplinary studies to understand the interactions within the human-environment system on aging processes. Todd et al.\cite{todd2020new} discuss the opportunities and challenges presented by the use of routine health and social care data for aging research, advocating for the development of standardized, validated algorithms to maximize the potential of routine data in understanding aging. Liu et al.\cite{aging2021aging} introduce the Aging Atlas, a multi-omics database aimed at providing a comprehensive profile of aging through the integration of various high-throughput 'omics' datasets. This database facilitates the exploration of the molecular profile and regulatory status of gene expression during aging. Lastly, Sturm et al. \cite{sturm2022multi} report on a multi-omic longitudinal dataset for studying aging in human fibroblasts, providing a resource for connecting mechanistic processes of aging with descriptive characterizations.

These studies illustrate the burgeoning role of data science in aging research, offering new methodologies for analyzing complex datasets and generating insights into the biological, environmental, and social determinants of aging. Through the application of big data analytics, machine learning, and multi-omics approaches, researchers are beginning to unravel the intricacies of aging, paving the way for interventions that could improve healthspan and quality of life for older adults.

\subsection{Deep Learning for Healthcare}
Deep learning is revolutionizing healthcare data science, offering transformative solutions for aging research, disease diagnosis, prognosis, and treatment optimization \cite{li2024research,lai2024language,tang2024zerothorder}. Studies such as those on MRI-based brain health evaluation \cite{bashyam2020mri}, biological age estimation \cite{rudnicka2020world} and early Alzheimer's detection \cite{Guo2020Resting}, demonstrate deep learning's capability to handle diverse and complex datasets, improving accuracy and efficiency in healthcare diagnostics and patient care. More specifically, deep representation learning in healthcare, particularly focusing on aging populations, leverages advanced deep learning techniques to gain insights from complex, high-dimensional data. The prowess in unraveling complex patterns within high-dimensional data is epitomized in studies by Moon et al. \cite{Moon2023Development} to discern latent biological aging markers in relation to morbidity and mortality, leveraging health check-up data enriched with crucial outcome information. This approach marks a significant leap in predicting health trajectories, demonstrating superior discriminability compared to traditional models. Similarly, DeepBrainNet \cite{bashyam2020mri} utilizes deep networks to generate robust brain-age estimates from MRI scans, offering a novel lens to assess brain health across the lifespan. The potential for early detection of Alzheimer’s disease through non-invasive means, as discussed by Ng et al. \cite{Ng2021Artificial}, highlights deep learning's role in revolutionizing diagnostics through retinal imaging, a less costly and invasive method pivotal for the aging population. Additionally, the systematic review by Si et al. \cite{Si2020Deep} underscores the transformative impact of deep representation learning on patient care management, disease diagnosis, and the advancement of personalized medicine through the analysis of electronic health records (EHRs). Collectively, these studies embody the potential of deep learning in healthcare, offering innovative solutions for patient stratification, disease prediction, and the management of aging populations, thus paving the way for a new era of medical interventions and healthcare delivery.

\subsection{HRS data Modeling}
The Health and Retirement Study (HRS) serves as a foundational dataset for understanding the multifaceted aspects of aging, retirement, and health dynamics in the United States. Recent research leveraging HRS data has provided critical insights into the drivers of rising mortality rates \cite{Monnat2022Enhancing}, cross-cultural measurement invariance in mood assessment \cite{Herrera2021Measurement}, and the significance of contextual data augmentation for enhancing research capabilities \cite{Dick2022The}. Studies by Mullen \cite{Mullen2022Using} have explored the impact of working conditions on later-life health and labor supply decisions, reflecting broader workforce trends. Additionally, the validation of self-reported cancer diagnoses against Medicare claims \cite{Mullins2021Validation} underscores the reliability and utility of HRS self-reported health data for population-based research. However, challenges remain, including ensuring data representativeness, addressing the complexity of modeling aging processes, and overcoming methodological hurdles related to measurement invariance and data linkage. These studies highlight the HRS's critical role in informing policy and intervention strategies, while also pointing to the need for continuous data enhancement and methodology.

\subsection{Multi-functional Degradation Modeling}
Recent advancements in degradation modeling, particularly in healthcare, have emphasized the importance of multi-functional or multi-source heterogeneity in predicting the failure times and trajectories of complex systems. Xiao et al. \cite{Xiao2020Degradation} developed a random effects Wiener process model to handle multi-source heterogeneity in degradation data, demonstrating improved model fitting and prediction accuracy by acknowledging variations in systems and measurement errors. Wang et al. \cite{Wang2021A} introduced a novel data fusion method based on deep learning for constructing a composite health index (HI) from multisensor data, significantly enhancing the accuracy of remaining useful life (RUL) predictions in complex systems. Gupta et al. \cite{gupta2021development} focused on developing a multivariable logistic regression model for predicting clinical deterioration among hospitalized adults with COVID-19, showcasing the potential of multi-functional modeling in healthcare settings.

These studies highlight the evolving landscape of degradation modeling and underscore the focus on integrating various data sources and leveraging advanced modeling techniques to capture the intricate dynamics of degradation process. Meanwhile, unlike existing regression-based methods, using deep representation learning for multi-functional degradation modeling could potentially mark significant strides toward more accurate and personalized healthcare services, which aligns with the focus of this work.

\section{Method}
In this section, we present our modeling framework on multi-functional degradation data using deep representation learning and quantify the latent heterogeneity through clustering techniques. First, we outline the general framework of deep representation learning to handle trajectory data. Second, we illustrate the multi-functional modeling framework using Long Short-Term Memory (LSTM) networks \cite{dai2023addressing}. Third, we integrate clustering techniques into representation learning process. 
\subsection{Deep Representation Learning For Trajectory}
Let \(X = \{X_1, X_2, \ldots, X_T\}\) be a sequence representing a trajectory, where each \(X_t \in \mathbb{R}^n\) is a vector describing the state at time \(t\), and \(T\) is the length of the trajectory. A deep learning model, such as a recurrent neural network (RNN) or Long Short-Term Memory (LSTM) network, is used to process the sequence data, which can be defined as 
\begin{equation}
    h_t = f(X_t, h_{t-1}; \Theta),
\end{equation}

where \(h_t\) is the hidden state at time \(t\), \(f\) is a non-linear function implemented by the network, and \(\Theta\) represents the model parameters. The goal is to learn a representation \(Z\) that captures the essential features of the trajectory. This can be formulated as an optimization problem:

\begin{equation}
    \min_{\Theta} L(Y, \hat{Y}(Z; \Theta)),
\end{equation}

where \(L\) is a loss function measuring the discrepancy between the true labels \(Y\) and the predicted labels \(\hat{Y}\), which are a function of the learned representation \(Z\).

\subsection{Multi-functional Degradation Modeling}
Multi-functional degradation modeling aims to predict the future state of a system experiencing degradation across multiple functions or dimensions. Long Short-Term Memory (LSTM) networks, a type of recurrent neural network, are particularly suited for this task due to their ability to model complex temporal dependencies. Consider a system where each sample has a trajectory represented by a sequence of multi-dimensional vectors \(X = \{X_1, X_2, \ldots, X_T\}\), with each \(X_t \in \mathbb{R}^n\) for \(n\) dimension of features at time \(t\), and static features \(S \in \mathbb{R}^m\) that are constant for each trajectory. The LSTM model processes the sequence data, integrating time-dependent and static features to predict the system's future degradation state. The LSTM is defined by the following equations:

\begin{itemize}
    \item Forget Gate:
    \[f_t = \sigma(W_f \cdot [h_{t-1}, X_t, S] + b_f)\]
    
    \item Input Gate:
    \[i_t = \sigma(W_i \cdot [h_{t-1}, X_t, S] + b_i)\]
    \[\tilde{C}_t = \tanh(W_C \cdot [h_{t-1}, X_t, S] + b_C)\]
    
    \item Update Cell State:
    \[C_t = f_t \ast C_{t-1} + i_t \ast \tilde{C}_t\]
    
    \item Output Gate:
    \[o_t = \sigma(W_o \cdot [h_{t-1}, X_t, S] + b_o)\]
    \[h_t = o_t \ast \tanh(C_t)\]
\end{itemize}

where \(\sigma\) is the sigmoid function, \(W\) and \(b\) are the weights and biases, respectively, and \([h_{t-1}, X_t, S]\) denotes the concatenation of the previous hidden state, the current input, and the static features.

The final state \(h_T\) is used to predict the next degradation condition \(Y_{T+1}\) through a fully connected layer:

\[Y_{T+1} = \phi(W_y \cdot h_T + b_y)\]

\begin{figure}[h!]
	\centering
	\includegraphics[width=\linewidth]{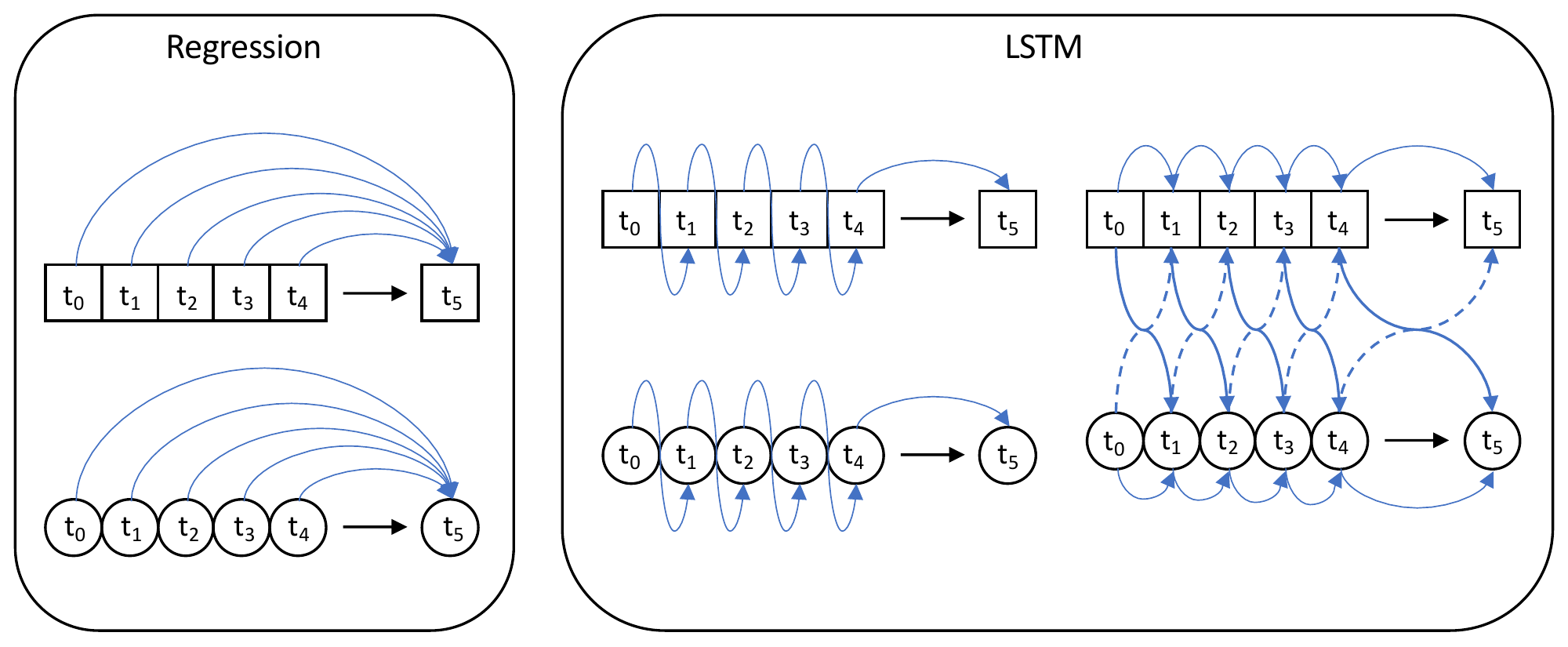}
		\caption{\small{Baseline vs Proposed Models}}  
	\label{model}
\end{figure}

The network is trained by minimizing a loss function \(L\), such as the Mean Squared Error (MSE) for continuous degradation measures:

\[L = \frac{1}{N} \sum_{t=1}^{N} (Y_{t+1} - \hat{Y}_{t+1})^2\]
Figure~\ref{model} illustrates the difference between the traditional regression model and the proposed multi-functional modeling framework with LSTM.

\begin{algorithm}
\small
\caption{Modeling Multi-Dimensional Trajectory Data with LSTM}
\begin{algorithmic}[1]

\State \textbf{Input:} Sequence of multi-dimensional trajectory data $\{X_1, X_2, \ldots, X_T\}$, static features $S$
\State \textbf{Output:} Predicted next state $Y_{T+1}$

\Procedure{InitializeLSTMModel}{}
    \State Initialize LSTM weights $W$, biases $b$
\EndProcedure

\Procedure{PreprocessData}{$X, S$}
    \State Normalize $X$ and $S$
    \State Split data into training and testing sets
\EndProcedure

\Procedure{TrainLSTM}{$X, S, Y$}
    \For{$epoch = 1$ to $N_{epochs}$}
        \For{$t = 1$ to $T-1$}
            \State $f_t = \sigma(W_f \cdot [h_{t-1}, X_t, S] + b_f)$
            \State $i_t = \sigma(W_i \cdot [h_{t-1}, X_t, S] + b_i)$
            \State $\tilde{C}_t = \tanh(W_C \cdot [h_{t-1}, X_t, S] + b_C)$
            \State $C_t = f_t \ast C_{t-1} + i_t \ast \tilde{C}_t$
            \State $o_t = \sigma(W_o \cdot [h_{t-1}, X_t, S] + b_o)$
            \State $h_t = o_t \ast \tanh(C_t)$
        \EndFor
        \State $Y_{T+1} = \phi(W_y \cdot h_T + b_y)$
        \State Compute loss $L(Y_{T+1}, \hat{Y}_{T+1})$
        \State Update $W$, $b$ using backpropagation
    \EndFor
\EndProcedure

\Procedure{PredictNextState}{$X, S$}
    \State Execute \Call{TrainLSTM}{$X, S, Y$} to get trained model
    \State Use $h_T$ from last time step to predict $Y_{T+1}$
    \State \textbf{return} $Y_{T+1}$
\EndProcedure

\end{algorithmic}
\end{algorithm}

\subsection{Clustering}
Upon obtaining the latent embeddings, we apply clustering techniques to group the embeddings into distinct clusters. Each cluster represents a subpopulation with similar characteristics in their trajectory patterns. The choice of clustering algorithm (e.g., K-means, DBSCAN, or hierarchical clustering) depends on the specific characteristics of the data and the research objectives.

\begin{algorithm}
\small
\caption{K-means Clustering on Latent Embeddings}
\begin{algorithmic}[1]

\State \textbf{Input:} Latent embeddings $Z = \{z_1, z_2, \ldots, z_M\}$, Number of clusters $K$
\State \textbf{Output:} Cluster assignments for each embedding

\Procedure{InitializeCentroids}{$E, K$}
    \State Randomly select $K$ embeddings from $E$ as the initial centroids $C = \{c_1, c_2, \ldots, c_K\}$
\EndProcedure

\Procedure{AssignClusters}{$E, C$}
    \Repeat
        \For{each embedding $e_i \in E$}
            \State Assign $e_i$ to the nearest centroid in $C$
        \EndFor
        \For{$k = 1$ to $K$}
            \State Update centroid $c_k$ as the mean of all embeddings assigned to cluster $k$
        \EndFor
    \Until{centroids $C$ do not change}
\EndProcedure

\State \textbf{return} Cluster assignments

\end{algorithmic}
\end{algorithm}

\section{Case study}\label{Casestudy}
In the case study, the real-world survey data from the Health and Retirement Study (HRS) is used to demonstrate the effectiveness of the proposed model in predicting the longitudinal degradation performances. The multidimensional heterogeneity in both physical and cognitive degradations can be quantified by automatically determining the number of subpopulations. The prediction performance of the proposed work can outperform the alternative models without heterogeneity quantification. The impacts of the degradation modeling on healthcare utilizations, including hospitals, nursing homes, etc., are explicitly evaluated. The final part of the case study further explores which factors may have resulted in the heterogeneities among subpopulations and how they can be associated with healthcare utilizations for long-term care management. 

\subsection{Data Description}
The data comes from the HRS data products \footnote{https://hrs.isr.umich.edu/data-products} \cite{chien2015rand, bugliari2023rand}, which is a panel study of aging people in America over 25 years and is the leading resource for data on aging. The HRS was launched in 1992 at the University of Michigan's Institute for Social Research to provide data for research on the aging of the population over age 50. It is also under the support of the National Institute on Aging (NIA) and the Social Security Administration (SSA). The study provides broad multidisciplinary longitudinal measurements such as health, health services utilization, income and wealth, family connections, etc. Since 1992, the study has conducted face-to-face core interviews every two years as a wave and collects detailed information on the well-being of elderly people. Currently, the core interview data are available for twelve waves from 1992 to 2014, with a total of over 20000 participants. With the rich longitudinal data, it is possible to model the degradation performance of the elderly population in order to provide better proactive healthcare preparedness and health service utilization management. 

\subsection{Data Analytics}
In the real case study, a retrospective participant cohort was selected with the following criteria: 1) age 65 or older; 2) continuous core interviews occurred from 1998 to 2012 for 8 waves. The sample data collects detailed information of 1699 participants, including their health degradation measurements, demographics, and health services utilization. \Cref{table-profile} provides the list of measurements and features considered in the case study. The joint consideration of both physical and cognitive degradations can better describe the overall health-changing patterns of the participants. The physical degradation is measured by the index of Activities of Daily Living (ADL), which combines the measures of multiple physical activities (e.g., walking, dressing, climbing, eating, etc.) in the elderly population's daily life. The cognitive degradation is measured by the index of Cognition (COG), which measures the cognitive functions such as counting, word recalling, naming, etc. Both indexes are integers ranging from 0 to 30. Higher index value may indicate lower physical performance for ADL or higher cognitive performance for COG. Demographical information can be useful in identifying the significant factors that may impact the health degradation of participants. The demographics include personal identity characteristics (e.g., age, gender, race, etc.), living habits (e.g., smoking, drinking, etc.), and economic features (e.g., poverty, wealth, income, etc.). Managing health services utilization is affected by and can be beneficial from accurate prediction of health degradations. The measurements of health services utilization include features for both acute care (i.e., hospital) and long-term care (i.e., nursing home). Specifically, categorical features such as overnight hospital stay (OHS), and overnight nursing home stay (ONHS) describe whether the care services have been utilized during the interview wave of two years. Numerical features describe the utilization degrees of each service through the sum of all overnight stays (e.g, OHSs, ONHSs) and the sum of nights for all overnight stays (e.g., NoOHSs, NoONHSs).

\begin{table}[h!]\renewcommand{\arraystretch}{1}\addtolength{\tabcolsep}{-5pt}
	\centering
        \footnotesize
	\caption{{Descriptive statistics of the selected sample}}
	\vspace{-0.4cm}%Workaround to be conform with the .doc style. Only for table captions.
	\begin{center}
		\begin{tabularx}{1\linewidth}{ll}
			\hline
			Features  & Statistics\\
			\hline
			Number of samples & 1699\\
			Number of waves & 8\\
			\textbf{Degradation measurements}  & \\
			\quad Physical -- ADL (mean, SD)  & 3.5 (4.32)\\
			\quad Cognitive -- COG (mean, SD)  & 22.06 (4.76)\\
			\textbf{Demographics}  & \\
			\quad Age (mean, SD)  & 70.07 (4.47)\\
			\quad Gender - Female ($n$, \%) &1154 (67.90\%)\\
			\quad Gender - Male ($n$, \%) &545 (32.10\%)\\
			\quad Smoke - Never ($n$, \%) &907 (53.40\%)\\
			\quad Smoke - Always ($n$, \%) &677 (39.80\%)\\
			\quad Smoke - Ever ($n$, \%) &115 (6.80\%)\\
			\quad Drink - Light / Never ($n$, \%) &1353 (79.60\%)\\
			\quad Drink - Heavy ($n$, \%) &225 (13.20\%)\\
			\quad Drink - Always ($n$, \%) &121 (7.10\%)\\	
			\quad Married - Yes, with spouse ($n$, \%) &1118 (65.80\%)\\
			\quad Married - Yes, spouse absent ($n$, \%) &395 (23.30\%)\\
			\quad Married - No, partnered ($n$, \%) &186 (10.90\%)\\
			\quad Region - Northeast ($n$, \%) &276 (16.20\%)\\
			\quad Region - Midwest ($n$, \%) &492 (29.00\%)\\
			\quad Region - South ($n$, \%) &620 (36.50\%)\\
			\quad Region - West ($n$, \%) &310 (18.20\%)\\
			\quad Region - Other ($n$, \%) &1 (0.10\%)\\
			\quad Race - White / Caucasian ($n$, \%) &1400 (82.40\%)\\
			\quad Race - Black / African American ($n$, \%) &174 (10.20\%)\\
			\quad Race - Hispanic / Latino ($n$, \%) &99 (5.80\%)\\
			\quad Race - Other ($n$, \%) &26 (1.50\%)\\
			\quad Poverty -  Category 1, poorest ($n$, \%) &250 (14.70\%)\\
			\quad Poverty -  Category 2 ($n$, \%) &459 (27.00\%)\\
			\quad Poverty -  Category 3 ($n$, \%) &351 (20.70\%)\\
			\quad Poverty -  Category 4 ($n$, \%) &357 (21.00\%)\\
			\quad Poverty -  Category 5, richest ($n$, \%) &282 (16.60\%)\\
			\quad Poverty threshold (mean, SD)&333 (394.64)\\
			\quad Education years (mean, SD)&12.31 (3.06)\\
			\quad Wealth (mean, SD)&367K (711K)\\
			\quad Income (mean, SD)&41K (49K)\\	
			
			\textbf{Health services utilization}  & \\
			\quad OHS - Yes ($n$, \%) &3628 (26.70\%)\\
			\quad OHS - No ($n$, \%) &9964 (73.30\%)\\
			\quad ONHS - Yes ($n$, \%) &414 (3.00\%)\\
			\quad ONHS- No ($n$, \%) &13178 (97.00\%)\\
			\quad OHSs (mean, SD)&0.41 (1.08)\\
			\quad NoOHSs (mean, SD)&1.77 (5.8)\\
			\quad ONHSs (mean, SD)&0.04 (0.22)\\
			\quad NoONHSs (mean, SD)&1.78 (23.82)\\
			
			\hline
			\noalign{\vskip 0.1mm} 
			\multicolumn{2}{l}{\tabincell{l}{SD: standard deviation}}\\
		\end{tabularx}
	\end{center}
	\label{table-profile}
\end{table}

\begin{figure}[h!]
	\centering
	\begin{subfigure}[b]{0.49\linewidth}
		\centering
		\includegraphics[width=\linewidth]{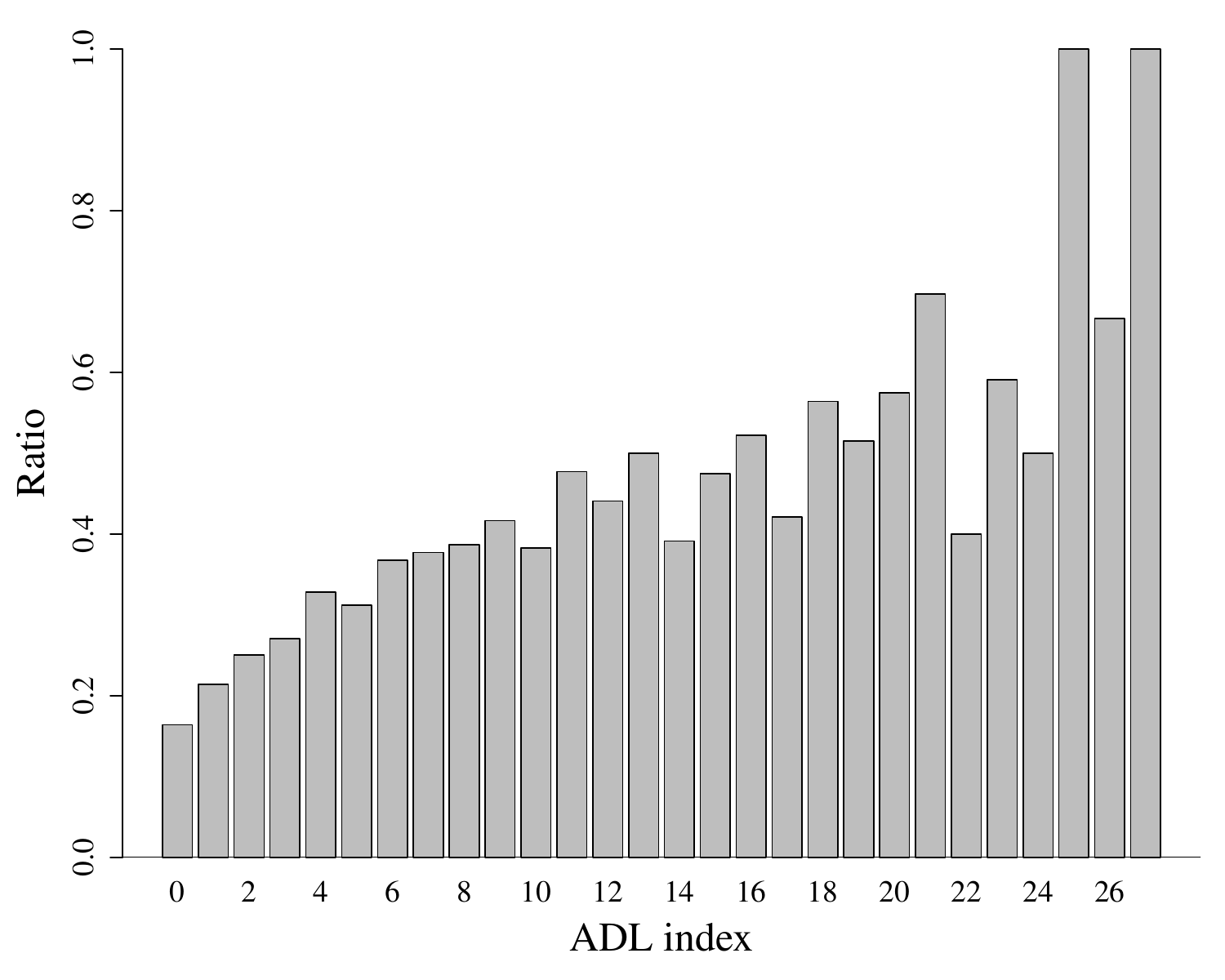}
		\caption{\small{ADL -- HS}}  
		\label{ADL_HOSP}
	\end{subfigure}
	\begin{subfigure}[b]{0.49\linewidth}
		\centering
		\includegraphics[width=\linewidth]{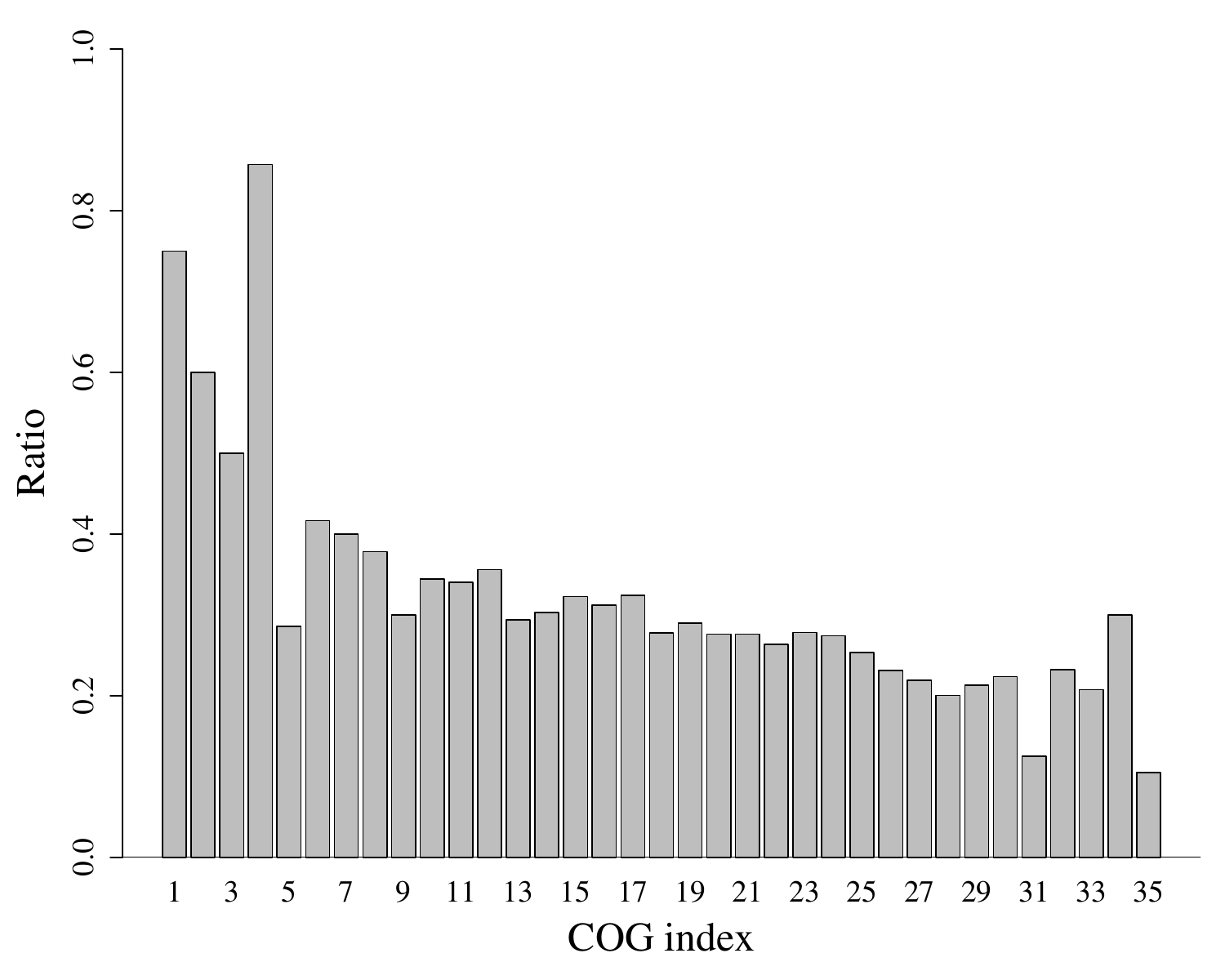}
		\caption{\small{COG -- HS} }
		\label{COG_HOSP}
	\end{subfigure}
	\begin{subfigure}[b]{0.49\linewidth}
		\centering
		\includegraphics[width=\linewidth]{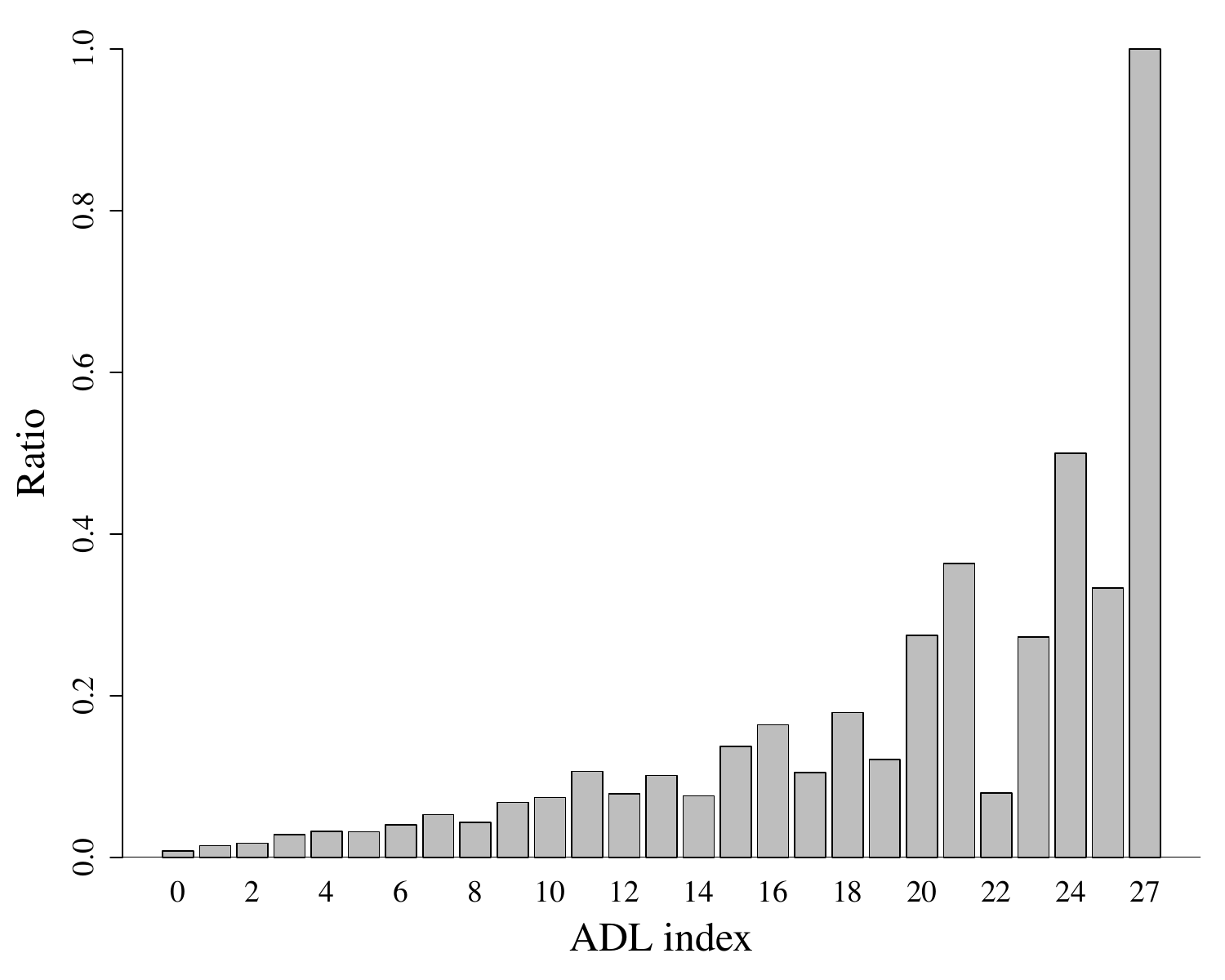}
		\caption{\small{ADL -- NHS}}  
		\label{ADL_NRSHOM}
	\end{subfigure}
	\begin{subfigure}[b]{0.49\linewidth}
		\centering
		\includegraphics[width=1\linewidth]{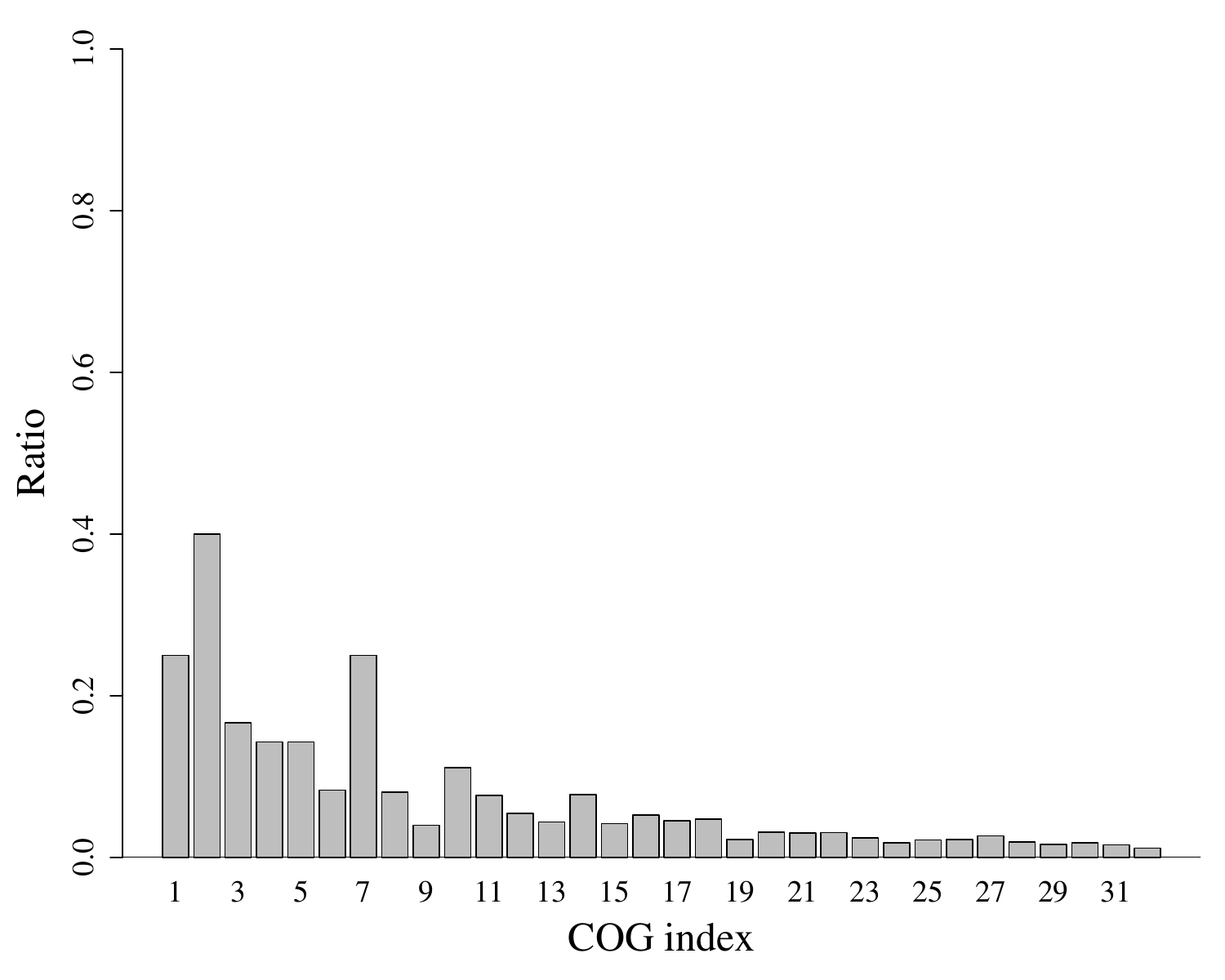}
		\caption{\small{COG -- NHS} }
		\label{COG_NRSHOM}
	\end{subfigure}
	\vspace{-2.2ex}
	\caption{\small{Occurrence Ratios of health services utilizations regarding health degradations in ADL and COG}}  
	\label{Utilization-profile}
\end{figure}

\Cref{Utilization-profile} shows the profiling of health services utilizations over both ADL and COG. As ADL increases, meaning the physical functionality is deteriorating, the ratios increase for both hospital and nursing home utilizations. Oppositely, lower COG, which means worse cognitive functionality, corresponds to higher services utilizations in hospital and nursing home. 

\subsection{Prediction Performance}
Table~\ref{tab: performance} presents a comparison of prediction performances between regression models and Long Short-Term Memory (LSTM) networks based on Mean Squared Error (MSE) for Activities of Daily Living (ADL) and Cognitive (COG) functions. The use of LSTM models yields a notable improvement in predictive accuracy, particularly when multi-dimensional data ("Multi") and additional features are considered in the modeling process.

For ADL, the best-performing LSTM model, which incorporates both multi-dimensional data and additional features, achieves an MSE of 1.12, a substantial enhancement over the regression models, where the lowest MSE reported is 1.20. This improvement is even more pronounced in the prediction of cognitive functions, with the MSE dropping to 0.02 in the optimal LSTM model from a minimum of 13.20 in the regression models.

The results demonstrate that LSTM models, especially when leveraging multi-dimensional trajectory data and a richer feature set, outperform traditional regression methods in predicting the progression of ADL and cognitive functions. This superior performance underscores the potential of LSTM networks in handling the complexities of healthcare data, offering significant advantages for predictive tasks in healthcare settings where precision is critical. The stark contrast in MSE for cognitive function predictions particularly highlights the LSTM's ability to capture intricate patterns and dependencies within the data, which are often missed by simpler regression techniques.
\begin{table}
    \small
    \centering
    \resizebox{0.47\textwidth}{!}{
    \begin{tabular}{c|c|c|cc}
			\hline
			% Method & Homogeneous& Target& MSE\\
                \multirow{2}{*}{Method}&\multirow{2}{*}{Multi}&\multirow{2}{*}{Features}&\multicolumn{2}{c}{MSE}\\
                &&&ADL&COG\\
                \hline
                Regression &  &  & 2.39 & 25.64\\
                Regression &x &  & 1.24 & 13.99\\
                Regression &  & x & 1.20 & 13.71\\
                Regression &x & x & 1.21 & 13.20\\
                % Regression & & & & 0 & 0 \\
                % Regression &x & & & 1.60 & 13.68\\
                % Regression & & & x & 0 & 0\\
                % Regression &x & & x & 1.54 & 13.72\\
                LSTM &  &  & 2.20 & 25.45\\
                LSTM &x &  & \textbf{1.12} & \underline{0.19}\\
                LSTM &  & x & 1.18 & 12.70\\
                LSTM &x & x & \underline{1.13} & \textbf{0.02}\\
                % LSTM & & & & 0 & 0\\
                % LSTM &x& & & 0 & 0 \\
                % LSTM & & & x & 0 & 0 \\
                % LSTM &x& & x & 0 & 0\\
			\hline 					
		\end{tabular}
    }
    \caption{\small Performance Comparison}
    \label{tab: performance}
    \vspace{-3mm} 
\end{table}

\subsection{Heterogeneous Clusters}
After training an LSTM model on multi-dimensional trajectory data, the resulting embeddings encapsulate the complex temporal dependencies of the dataset in a high-dimensional latent space. K-means clustering is then applied to these embeddings to identify distinct groups, or clusters, based on the similarity of their degradation patterns.

\begin{figure}[h!]
	\centering
	\includegraphics[width=\linewidth]{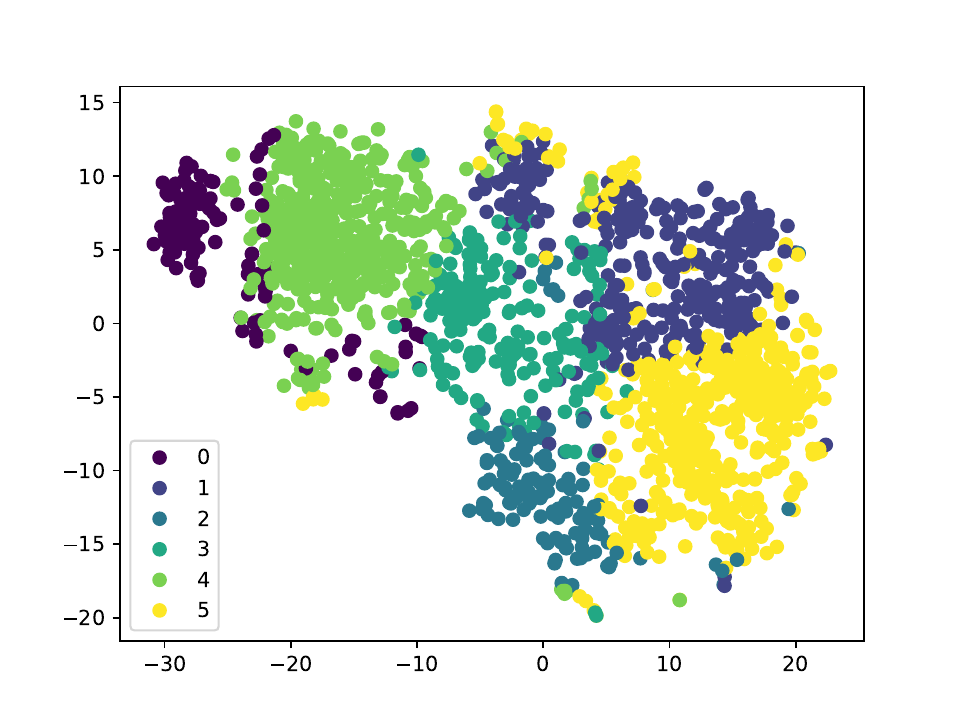}
		\caption{\small{Heterogeneous Clusters from multi-functional degradation embeddins}}  
	\label{tsne_Cluster}
  \vspace{-3mm}
\end{figure}

Once the clusters are defined, t-SNE comes into play as a powerful tool to visualize these high-dimensional clusters in a two-dimensional space, making it easier to interpret the results, as shown in Figure~\ref{tsne_Cluster}. The t-SNE algorithm effectively maps the high-dimensional data to a lower-dimensional space while preserving the local structure of the data, which means that embeddings close to each other in the high-dimensional space will likely appear close in the two-dimensional t-SNE plot.

The t-SNE plot provides a visual confirmation of the heterogeneity within the data, as clusters formed by K-means become visually discernible. This visualization can reveal the degree to which the LSTM model has been able to differentiate between various subpopulations, indicating the presence of distinct progression patterns or states of degradation within the data. It can also identify any outliers or anomalies that may not conform to the identified patterns.

\subsection{Decision Making Strategy}
\begin{figure}[h!]
	\centering
	\begin{subfigure}[b]{0.49\linewidth}
		\centering
		\includegraphics[width=\linewidth]{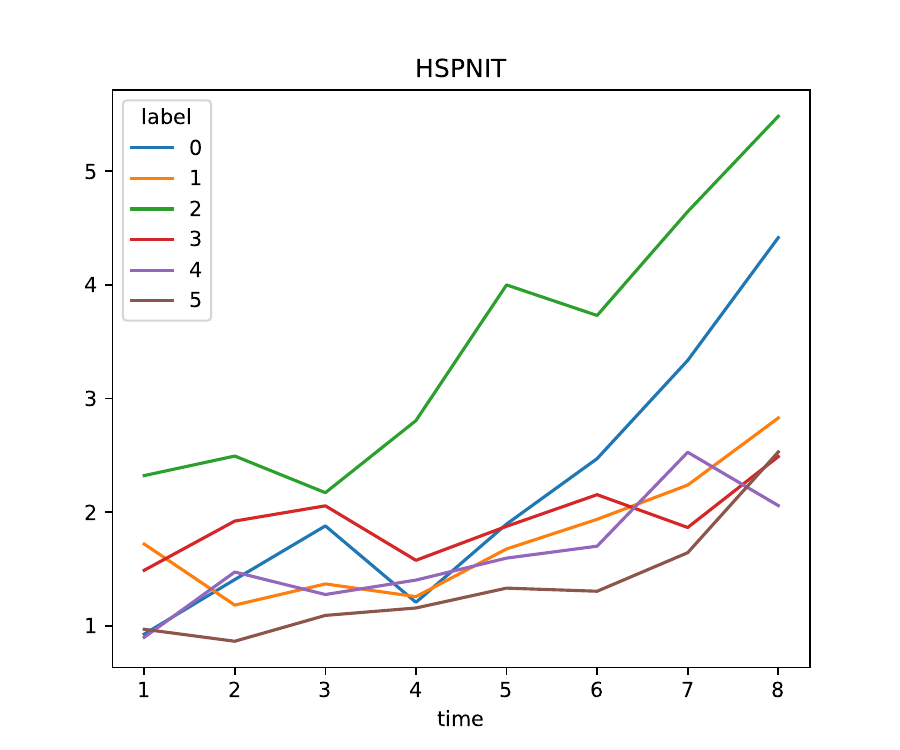}
		\caption{\small{HSPNIT}}  
		\label{HSPNIT}
	\end{subfigure}
	\begin{subfigure}[b]{0.49\linewidth}
		\centering
		\includegraphics[width=\linewidth]{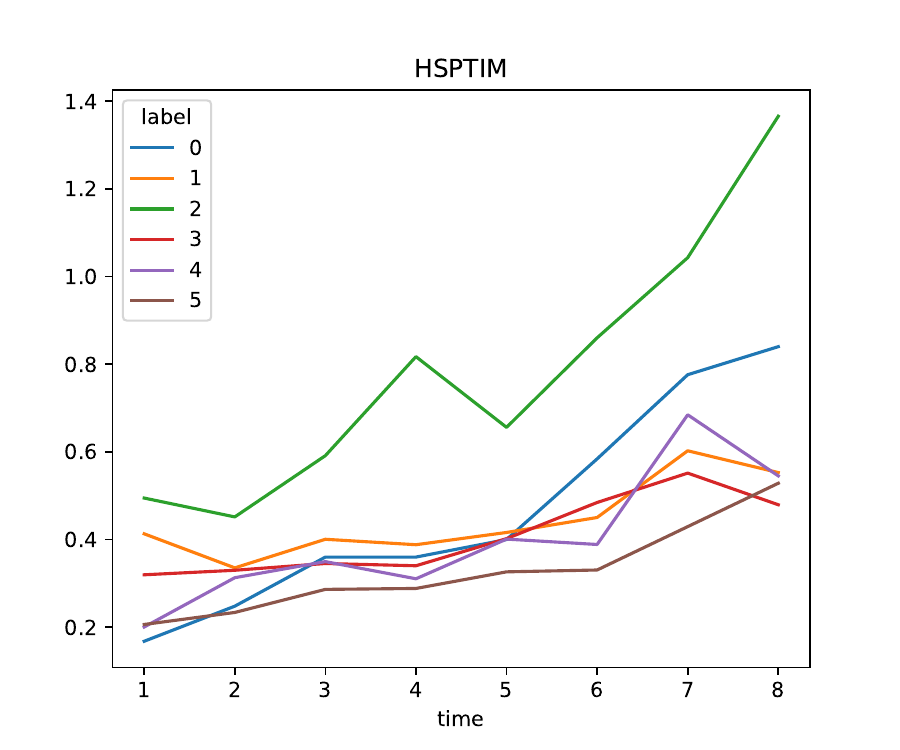}
		\caption{\small{HSPTIM} }
		\label{HSPTIM}
	\end{subfigure}
	\begin{subfigure}[b]{0.49\linewidth}
		\centering
		\includegraphics[width=\linewidth]{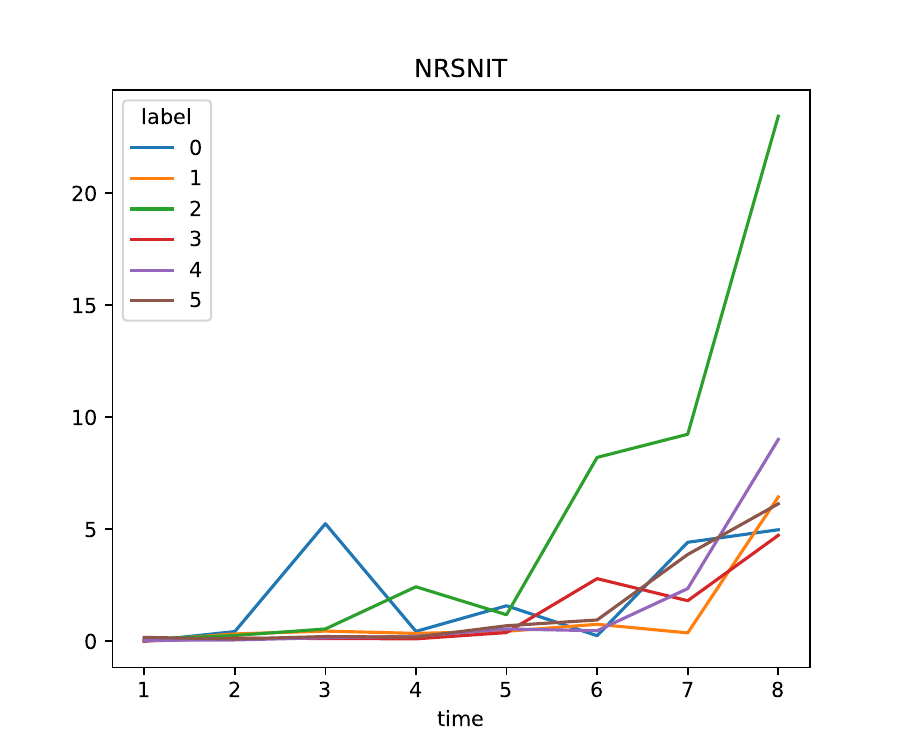}
		\caption{\small{NRSNIT}}  
		\label{NRSNIT}
	\end{subfigure}
	\begin{subfigure}[b]{0.49\linewidth}
		\centering
		\includegraphics[width=1\linewidth]{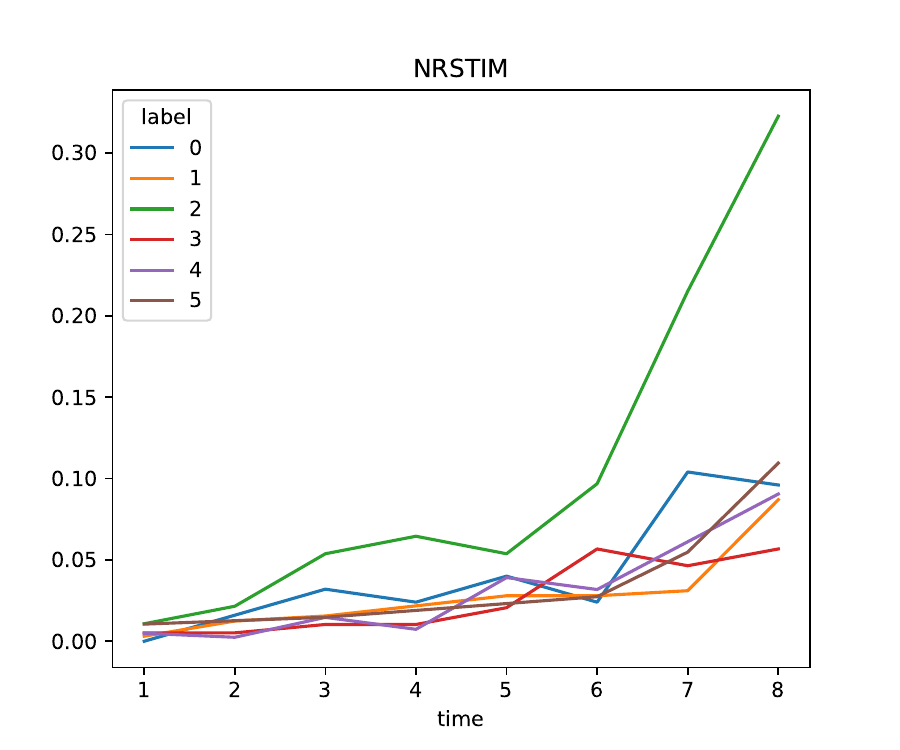}
		\caption{\small{NRSTIM} }
		\label{NRSTIM}
	\end{subfigure}
	\vspace{-2.2ex}
	\caption{\small{Occurrence ratios of health service utilizations for heterogeneous degradation in ADL and COG}}  
	\label{Utilization-cluster}
\end{figure}

Leveraging the results from K-means clustering of LSTM-derived embeddings, we can effectively quantify the healthcare utilization patterns inherent to each identified heterogeneous cluster. This granular analysis permits a clearer understanding of how different subpopulations, characterized by unique multi-functional degradation trajectories, consume healthcare resources. By doing so, it becomes possible to anticipate healthcare demands more accurately, facilitating improved planning and management of healthcare services. For instance, Figure~\ref{Utilization-cluster} illustrates that the utilization of healthcare services has heterogeneous patterns for different clusters. By aligning healthcare delivery models with the diverse requirements of these clusters, healthcare providers can optimize resource allocation, personalize treatment plans, and, ultimately, enhance patient care outcomes. This targeted approach to healthcare utilization not only stands to improve the efficiency of healthcare systems but also supports the strategic development of healthcare policies tailored to the nuanced needs of the aging population.

\section{Conclusion}
In this study, we presented a pioneering methodology for modeling multi-functional degradation trajectories in healthcare, utilizing Long Short-Term Memory (LSTM) networks for deep representation learning of complex, temporal dynamics in multi-dimensional trajectory data. This methodology surpasses traditional univariate and homogeneous models by capturing the multifaceted nature of degradation and the diversity within aging populations, thus offering a nuanced understanding of individual and collective degradation pathways. The integration of dynamic trajectory data with static patient characteristics further enhances the model's predictive power, facilitating personalized healthcare interventions based on precise degradation profiles. Our approach's validation against real-world healthcare data underscores its potential to revolutionize decision-making processes in healthcare systems, allowing for optimized resource allocation and tailored care strategies that address the specific needs of distinct patient subgroups. By revealing the inherent heterogeneity of degradation patterns and their implications for healthcare delivery, our method sets a new standard for predictive analytics in healthcare, promising to improve care outcomes through more informed, data-driven decisions. This study not only demonstrates the feasibility and value of leveraging advanced machine learning techniques for healthcare analytics but also highlights the critical need for methodologies that accommodate the complexity and diversity of patient data in the pursuit of enhanced healthcare planning and intervention.

\bibliographystyle{plain}
\bibliography{SDM2024}

\end{document}